\pdfoutput=1

\documentclass[11pt]{article}

\usepackage[]{EMNLP2022}

\usepackage{times}
\usepackage{latexsym}
\usepackage{graphicx}
\usepackage{multirow}

\usepackage[T1]{fontenc}

\usepackage[utf8]{inputenc}

\usepackage{microtype}

\usepackage{inconsolata}

\title{Word Sense Induction with Hierarchical Clustering and Mutual Information Maximization}

\author{Hadi Abdine$^{1}$, Moussa Kamal Eddine$^{1}$, Michalis Vazirgiannis$^{1,2}$, Davide Buscaldi$^{3}$ \\
        $^{1}$École Polytechnique, $^{2}$AUEB, $^{3}$Université Sorbonne Paris Nord }

\begin{document}
\maketitle
\begin{abstract}

Word sense induction (WSI) is a difficult problem in natural language processing that involves the unsupervised automatic detection of a word's senses (i.e. meanings).
Recent work achieves significant results on the WSI task by pre-training a language model that can exclusively disambiguate word senses, whereas others employ previously pre-trained language models in conjunction with additional strategies to induce senses. 
In this paper, we propose a novel unsupervised method based on hierarchical clustering and invariant information clustering (IIC). 
The IIC is used to train a small model to optimize the mutual information between two vector representations of a target word occurring in a pair of synthetic paraphrases. This model is later used in inference mode to extract a higher quality vector representation to be used in the hierarchical clustering.  
 We evaluate our method on two WSI tasks and in two distinct clustering configurations (fixed and dynamic number of clusters). We empirically demonstrate that, in certain cases, our approach outperforms prior WSI state-of-the-art methods, while in others, it achieves a competitive performance. 
\end{abstract}

\section{Introduction}

\begin{figure*}[t!]
    \centering
    \includegraphics[width=13.3cm]{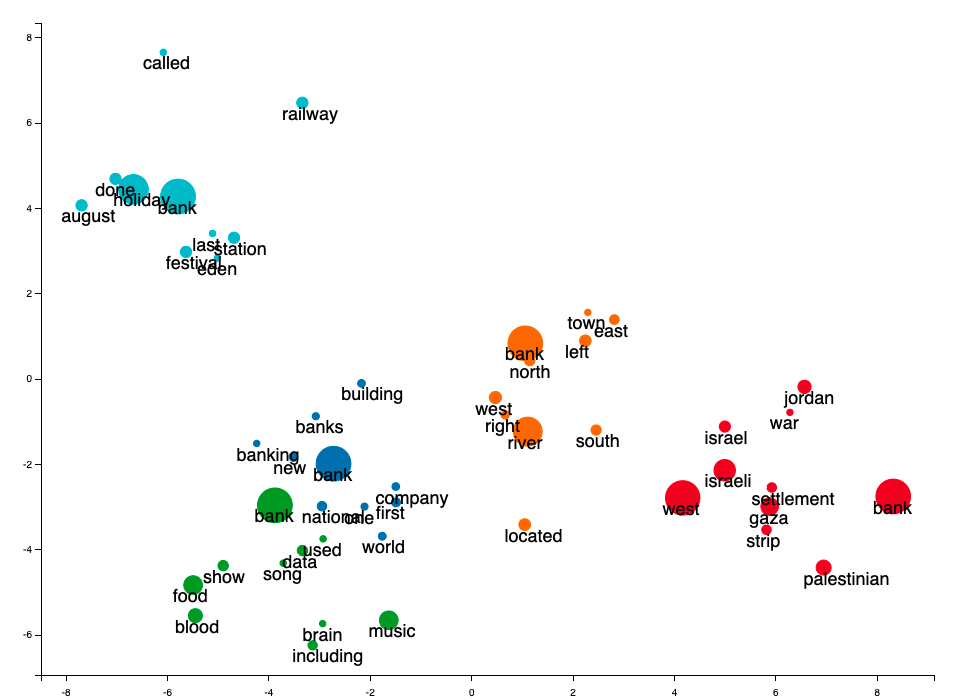}
    \caption{An illustration represents the different sense-based clusters of the word \textbf{bank} with the most frequent words used in the corresponding contexts. These clusters are obtained using agglomerative clustering on a set of RoBERTa vectors of the word \textbf{bank} extracted from 3000 sentences collected from Wikipedia. The centre of each cluster is the 2D PCA vector of the average 'bank' vectors of the cluster. The size of the points is proportional to the frequency of its appearance in the context of each sense-based cluster.}
    \label{fig:bank_clusters}
\end{figure*}

The automatic identification of a word's senses is an open problem in natural language processing, known as "word sense induction" (WSI). It is clear that the task of word sense induction is closely related to the task of word sense disambiguation (WSD) relying on a predefined sense inventory (i.e. WordNet \citep{wn1, wn2, wn3}) and aiming to solve the word's ambiguity in context. In WSI, given a target word, we focus on clustering a collection of sentences using this word according to its senses. For example, figure \ref{fig:bank_clusters} shows the different clusters obtained by using RoBERTa$_{LARGE}$ \citep{liu2019roberta} of 3000 sentences that contain the word bank collected from Wikipedia. We can see five different clusters where the corresponding centroids represent the 2D PCA projection of the average contextual word vectors of the word bank. The clusters are obtained using the agglomerative clustering with cosine affinity and average linkage.
Word senses are more beneficial than simple word forms for a variety of tasks including Information Retrieval, Machine Translation and others \citep{2002}. The former are typically represented as a fixed list of definitions from a manually constructed lexical database. However, lexical databases are missing important domain-specific senses. For example, these databases often lack explicit semantic or contextual links between concepts and definitions \citep{2001}. Hand-crafted lexical databases also frequently fail to convey the precise meaning of a target word in a specific context \citep{article2004}. In order to address these issues, WSI intends to learn in an unsupervised manner the various meanings of a given word.\\
This paper includes the following contributions:\\
1) We propose a new unsupervised method using contextual word embeddings (i.e. RoBERTa, BERT and DeBERTa \citep{he2021DeBERTa}) that are being updated with more sense-related information by maximizing the mutual information between two instances of the same cluster. To achieve that, we generate a randomly perturbated replicate of the given sentence while preserving its meaning. Thus, we extract different word representations of the same target with two similar contexts. This method presents competitive results on WSI tasks.\\
2) We apply for the first time a method to compute a dynamic number of senses for each word. We rely on a recent word polysemy score function \cite{xypolopoulos2020unsupervised}.\\
3) We study the sense information per hidden layer for four different pretrained language models. We share, for all models, the layers with the best performance on sense-related tasks.

\section{Related Work}
Previous works on WSI use generative statistical models to solve this task. Mainly, they approach this task as a topic modeling problem using Latent Dirichlet Allocation (LDA) \citep{lau2013, chang-etal-2014-inducing, goyal-hovy-2014-unsupervised, wang2015, komninos2016}. AutoSense \cite{amplayo2018autosense}, one of the most recent best-performing LDA methods, is based on two principles: First, senses are represented as a distribution over topics. Second, the model generates a pair composed of the target word and its neighboring word, thus seperating the topic distributions into fine-grained senses based on lexical semantics. 
AutoSense throws away the garbage senses by removing topics distributions that don't belong to any instance. Furthermore, it adds new ones according to the generated (target, neighbor) pairs which means that fixing the number of senses by the model is not required. While most of the WSI methods fix the number of clusters for all the words, in our work we explore two setups for the number of clusters, fixed and dynamic.
Other works \citep{song2016sense, correa2018} use the static word embedding Word2Vec \citep{mikolov2013efficient} to get the representations of polysemous words before applying the clustering method.\\
After the emergence of contextual word Embeddings, pretrained language models such as ELMo \citep{peters2018deep} (based on BiLSTM) and BERT \citep{devlin2019bert} (based on the transformers) \citep{vaswani2017attention} are used with additional techniques to induce senses of a target word. \citep{amrami-goldberg-2018-word} and \citep{amrami2019better} use consecutively ELMo and BERT$_{LARGE}$ to predict probable substitutes for the target words. Next, it gives each instance $k$ representatives where each one contains multiple possible substitutes drawn randomly from the word distribution predicted by the language model. Each representative is a vector conducted from TF-IDF. Following, the representatives are clustered using the agglomerative clustering where the number of clusters is fixed to 7. Finally, each instance will be assigned to one or multiple clusters according to the corresponding cluster of each of its representatives. Instead of using the word substitutes approach, our work uses the contextual word embedding extracted from pretrained language models.\\
PolyLM \citep{ansell-etal-2021-polylm} is one of the most recent techniques for word sense induction that uses a MLM (Masked Language Model) to induce senses. PolyLM took a novel approach to the problem of learning word senses. It uses the transformer architecture to predict eight probabilities for each word, where each probability represents the probability of a word to be assigned to one of eight different senses. It is built on two assumptions: the chance of a word being predicted in a masked place is proportional to the total of its distinct senses, and for a particular context, one of the word's senses is more likely to be used. The model has the drawback of assuming the same fixed number of senses for all words.

\section{Method}
Our method consists of four main steps: 
First, we form a dataset of pair of sentences. To guarantee the fact that the target word in one pair belongs to the same cluster-based sense, each sentence is joined with a randomly perturbed replicate. Further details can be found in \ref{perturbation}.
Second, we extract the pairwise hidden state representations of the target word from one of the layers of a pretrained language model. Hence, for each pair of sentences, we have now a pair of vectors.
Mainly, we will evaluate this method on  RoBERTa$_{LARGE}$, BERT$_{LARGE}$ and DeBERTa$^{mnli}_{XLARGE}$.
Third, we train a MIM (Mutual Information Maximization) model where: (1) Considering an instance of the pairwise hidden state representations, the network is trained with an objective function maximizing the mutual information and minimizing the match loss between the output of the two vectors. (2) The best instance of the model is chosen according to the least loss on the pre-defined test set. (3) The original vectors of the test set are now replaced with the output of the first layer as the new vectors for the target word.
Fourth, for each target word in the evaluation datasets, we apply the agglomerative clustering method on the new outputs to obtain our clustering solution. To choose the pre-defined number of clusters, we follow two approaches: (i) Fix the number of senses (clusters) to 7 as in \citet{amrami-goldberg-2018-word, amrami2019better} and (ii) Use a dynamic number of clusters based on the polysemy score \citep{xypolopoulos2020unsupervised} of each target word.\\
The main steps are detailed in the following subsections.
\begin{figure}[h!]
    \includegraphics[width=7.9cm]{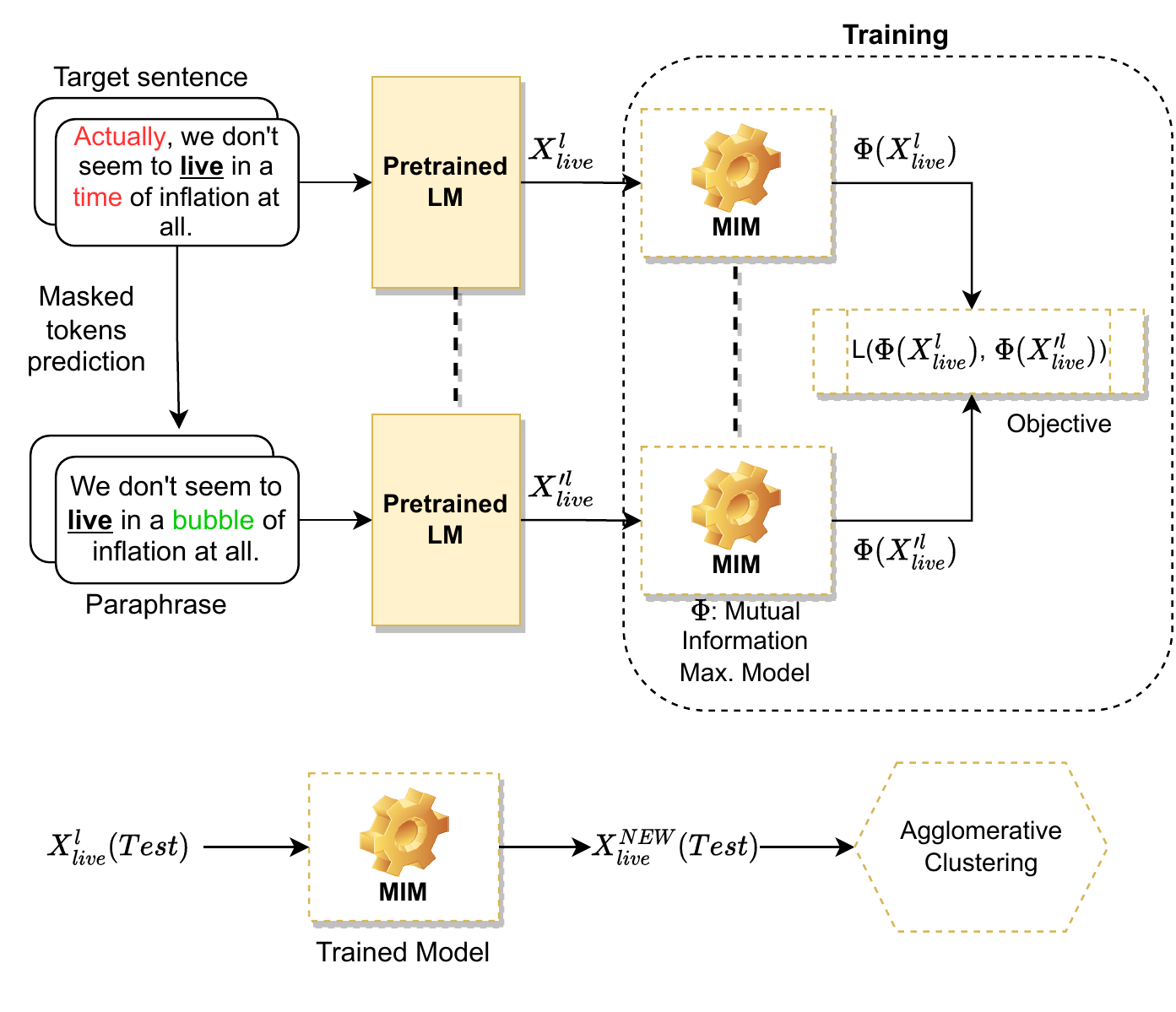}
    \caption{The pipeline of our method: For the word "live" chosen as target, a list of sentences is provided. BART is used to generate their corresponding paraphrases. The hidden representation X$_{live}^{l}$ of the target word is extracted from the layer \textbf{l} of a pretrained language model. Dashed line denotes shared parameters.}
    \label{fig:method}
\end{figure}
\subsection{Dataset Setup}
\label{perturbation}
\paragraph{BART} \citep{lewis-etal-2020-bart} is a denoising autoencoder for pretraining sequence-to-sequence models. It is trained by learning a model to rebuild a corrupted version of the original sentences using an arbitrary noising function. It is based on a standard Tranformer-based neural machine translation architecture which can be seen as a generalization of BERT (due to the bidirectional encoder), GPT \citep{Radford2018ImprovingLU} (with the left-to-right decoder), and other recent pretraining schemes. BART can be used also as a generative model given an input i.e. sentence completion, translation, summarization, etc..


\paragraph{Generating randomly perturbated replicates} In order to apply our method on the text input, we need to create a pair of sentences where the target word has the same sense. To fulfil this, a function is needed to introduce random perturbations to the input sentence while preserving the meaning. The sentence and it's perturbated version are keeping the same sense of the target lemma. Thus, we can generate a pair of sentences that belong to the same cluster.
First, we masked 40\% of the original sentence while preventing -in most cases- masking the target word. %
Second, we predicted the masked tokens using $BART_{BASE}$ with a beam size of one. 

\subsection{Vectors Extraction}\label{bertvectors}
The train set is used to train the parameters of a small network while the test set is used to perform the induction of the senses.
Using the best layer of each of the following transformer-based models: BERT$_{LARGE}$, RoBERTa$_{LARGE}$ and DeBERTa$_{XLARGE}$, we extracted representations of the target word from the different train and test instances. The best layer for each pretrained language model is chosen according to the best performance on BERTScore \citep{bert-score} with WMT16 To-English Pearson\footnote{\url{https://docs.google.com/spreadsheets/d/1RKOVpselB98Nnh_EOC4A2BYn8_201tmPODpNWu4w7xI/edit?usp=sharing}}.\\
At this stage, if the target word is broken down into multiple tokens using WordPiece tokenizer being used in the pretrained language model, we computed the average vector of the corresponding word pieces.
Note that, while generating the perturbation on the input text using BART$_{BASE}$, there is a small probability that the paraphrase might not contain the target word. Thus, all the sentences in the training set with their corresponding paraphrases deprived of the target word are removed.

\subsection{Loss Function}
We seek to minimize a loss function L with two components, each of which is explained in the following:
\begin{equation}
    L = L_{IIC} + L_{M}
\end{equation}
\subsubsection{Invariant Information Clustering Loss}
Invariant information clustering IIC \citep{ji2019invariant} is a clustering objective that learns a neural network from scratch to perform unsupervised image classification and segmentation. The model learns to cluster unlabeled data based on maximizing the mutual information score between the unlabeled sample and a transformation of the input. Therefore, both the input and its corresponding transformation surely contain the same information and do belong to the same class/cluster. Maximizing the mutual information is robust to clustering degeneracy where a single cluster tends to dominate the predictions or some clusters tend to disappear as in k-means. Also, it helps to avoid noisy data from affecting the predictions by over-clustering. The objective function is as follows:
\begin{equation}
    max_{\Phi} I(\Phi (x), \Phi(x'))
\end{equation}
Where $\Phi$ is the classification neural network, x is the input, and x'=g(x) is the transformation (random perturbation of the input) of x (i.e. rotation, maximizing, minimizing, etc..). This is equivalent to maximizing the predictability of $\Phi(x)$ from $\Phi(x')$ and vice versa.
The mutual information function is defined by:
\begin{equation}
    I(X, Y) = \sum_{y\in Y}\sum_{x\in X}P_{X,Y}(x, y)log\frac{P_{X,Y}(x, y)}{P_{X}(x)P_{Y}(y)}
\end{equation}
The loss of invariant information clustering is therefore defined by:
\begin{equation}
    L_{IIC} = -I(\Phi (x), \Phi(x'))
\end{equation}
We adopt the IIC loss to the NLP domain by changing the nature of the random perturbation introduced to the input.
\subsubsection{Match Loss}
The output of the model's last layer might be the same for all the different train sentences in some cases.
To tackle this issue, we encourage the similarity between the last layer's outputs $\Phi(x)$ and $\Phi(x')$ by adding a match loss. This loss is proportional to the cosine similarity between the two outputs and it is inspired from \citep{ansell-etal-2021-polylm} with the following:
\begin{equation}
   L_{M} = -0.1 \sum\frac{\Phi (x)\cdot\Phi (x')}{\|\Phi (x)\|\|\Phi (x')\|} 
\end{equation}

\subsection{Sense Embedding: Getting New Word Vectors }
The architecture of our MIM model is very simple. It is formed of three projection layers with ReLU activation function. The final layer is equipped with the softmax function to get a probability distribution vector usable as an input to our loss. The hidden size of one linear layer is set to the double of RoBERTa$_{LARGE}$'s and BERT$_{LARGE}$'s hidden state size which is 1024. \\
For each target word, we train a model while providing the pairs of extracted representations belonging to the same cluster. In other terms, the target word's representations from the original sentence and the sentence with lexical perturbation respectively.\\
The training concerns 8 runs over 5 epochs with a batch size of 32 using Adam optimizer \citep{https://doi.org/10.48550/arxiv.1412.6980}. The learning rate starts with 2e-5 and then is reduced linearly to zero over the remaining training time. The best model results from the epoch minimizing the validation loss. The validation set is the pairs of sentences from the test dataset.\\
Once the training is complete, the hidden state representation of the first layer is extracted for each test word vector of the original sentence. Thus, the target word has a new projected representation.

\subsection{Clustering}
To cluster the instances into senses, we used the agglomerative clustering method. The same setup as in \citep{amrami-goldberg-2018-word, amrami2019better} is used along with cosine distance and average linkage.
To choose the number of clusters (senses) of each target word, we follow two approaches: (i) Fix the number of senses as in \citep{amrami-goldberg-2018-word, amrami2019better, ansell-etal-2021-polylm}. (ii) Use a dynamic number of clusters based on its polysemy score obtained using the unsupervised word polysemy quantification \citep{xypolopoulos2020unsupervised}. For the dynamic clustering, we use the best configuration in the paper with dimensionality D equal to 3 and a level L equal to 8.

\section{Evaluation}
\label{sec:evaluation}
Several competitions were organized to systematically evaluate various methods applied for WSI, including \textit{SemEval-2007 task 02} \citep{agirre-soroa-2007-semeval}, \textit{SemEval-2010 task 14} \citep{manandhar-klapaftis-2009-semeval} and \textit{SemEval-2013 task 13} \citep{jurgens-klapaftis-2013-semeval}. The two tasks of \textit{SemEval-2010} and \textit{SemEval-2013} are considered as the benchmark for WSI.
In this section, we publish and analyse the mean and standard deviation over 8 runs of the previously described model on the two mentioned tasks: \textit{SemEval-2010 task 14} and \textit{SemEval-2013 task 13}.
\subsection{SemEval-2010 task 14:}
On one hand, the primary objective of the \textit{SemEval-2010} WSI challenge is to compare unsupervised word-sense induction systems. It provides a mapping mechanism for evaluating WSI systems using the WSD dataset. The target word dataset consists of 100 tagged words, 50 nouns and 50 verbs extracted from OntoNet \citep{ontonet}. In the test set, each target word has around one hundred instances to be clustered. To learn its senses, a training set containing approximately 10,000 instances is provided for each target word. The training set is created using a semi-automatic web-based method. For each sense of the target word in WordNet \citep{wn1}, the query grabs all the sentences containing its corresponding stems and lemmas using Yahoo! search API. Each instance in the test dataset in this task is labeled with one sense only. \\
The performance in this task is measured with V-Measure \citep{vmeasure} (biased toward high number of clusters) and F-Score (biased toward low number of clusters). We report the overall performance (\textbf{AVG}) defined as the geometric mean of these two metrics.

\subsection{SemEval-2013 task 13:} 
On the other hand, \textit{SemEval-2013 task 13} is a task for evaluating Word Sense Induction and Disambiguation systems in a context where instances are tagged with many senses whose applicability is weighted accordingly (Fuzzy Setting). The task focuses on disambiguating senses for 50 target lemmas: 20 nouns, 20 verbs, and 10 adjectives. The ukWac corpus \citep{baroni2009} is provided as a training corpus. It contains large number of instances crawled from the web and can be filtered by lemma, POS tag and many more filters\footnote{\url{https://corpora.dipintra.it/public/run.cgi/first\_form}}. Test data are drawn from the Open American National Corpus \citep{ide-suderman-2004-american} across a variety of genres and from both the spoken and written portions of the corpus. \\
The performance in this task is measured with Fuzzy B-Cubed (F-BC)\citep{bagga-baldwin-1998-entity}. It is a generalized version of B-Cubed that deals with the fuzzy setting and Fuzzy Normalized Mutual Information (F-NMI). The latter is a generalized version of mutual information that deals with multi-sense annotation. We report as well the overall performance (\textbf{AVG}).

\begin{table*}[t]
 \centering
 \small
 \begin{tabular}{||c c c c c||} 
 \hline
 Model &  \# Clusters & V-Measure & F-score & AVG\\ [0.5ex] 
 \hline\hline
 RoBERTa$_{LARGE}^{17}$  & 7 & 39.8 & 67.18 & 51.71 \\
 RoBERTa$_{LARGE}^{17}$ (+MIM) & 7 & \textbf{44.83±1.08} & 67.74±0.78 & \textbf{55.1±0.88} \\
 RoBERTa$_{LARGE}^{17}$  & Dynamic & 37 & 67.42 & 49.94 \\
 RoBERTa$_{LARGE}^{17}$ (+MIM) & Dynamic & 44.3±0.79 & 68.63±0.42 & 55.14±0.57  \\
 BERT$_{LARGE}^{18}$ & 7 & 40.1 & 65.23 & 51.14 \\
 BERT$_{LARGE}^{18}$ (+MIM)& 7 & 40.34±0.88 & 64.75±1.08 & 51.11±0.93 \\
 BERT$_{LARGE}^{18}$ & Dynamic & 41.2 & 67.17 & 52.6 \\
 BERT$_{LARGE}^{18}$ (+MIM)& Dynamic & 42.13±0.45 & 67.39±0.37 & 53.28±0.35 \\
 DeBERTa$_{XLARGE}^{40}$ & 7 & 40.5 & 66.64 & 51.95 \\
 DeBERTa$_{XLARGE}^{40}$ (+MIM)& 7 & 40.23±0.83 & 67.05±0.76 & 51.93±0.67 \\
 DeBERTa$_{XLARGE}^{40}$ & Dynamic & 40.6 & 67.52 & 52.36 \\
 DeBERTa$_{XLARGE}^{40}$ (+MIM)& Dynamic & 40.59±1.23 & 68.05±0.76 & 52.49±1.1 \\
 \hline
 PolyLM$_{BASE}$ \citep{ansell-etal-2021-polylm} & 8 & 40.5 & 65.8 & 51.6 \\
 PolyLM$_{SMALL}$ \citep{ansell-etal-2021-polylm} & 8 & 35.7 & 65.6 & 48.4 \\
 BERT+DP \citep{amrami2019better} &7 & 40.4 & \textbf{71.3} & 53.6 \\
 AutoSense \citep{amplayo2018autosense} & Dynamic &9.8 & 61.7 & 24.59 \\ 
\hline
\end{tabular}
\caption{Comparison of WSI-specific techniques on SemEval 2010 task 14}
\label{tab:semeval2010}
\end{table*}
%
\begin{table*}[t]
 \centering
 \small
 \begin{tabular}{||c c c c c||} 
 \hline
 Method &  \# Clusters & F-BC & F-NMI & AVG\\ [0.5ex] 
 \hline\hline
 RoBERTa$_{LARGE}^{17}$  & 7 & 64.1 & 19.28 & 35.16 \\
 RoBERTa$_{LARGE}^{17}$ (+MIM) & 7 & 62.87±0.5 & 21.84±0.74 & 37.07±0.73 \\
 RoBERTa$_{LARGE}^{17}$  & Dynamic & 64.2 & 16.11 & 32.16 \\
 RoBERTa$_{LARGE}^{17}$ (+MIM) & Dynamic & 65.02±0.44 & 20.86±0.66 & 36.83±0.68 \\
 BERT$_{LARGE}^{18}$ & 7 & 62.4 & 21.58 & 36.7 \\
 BERT$_{LARGE}^{18}$ (+MIM)& 7 & 62.8±0.49 & 22.74±0.5 & 37.79±0.52 \\
 BERT$_{LARGE}^{18}$ & Dynamic & 64.81 & 20.86 & 36.77 \\
 BERT$_{LARGE}^{18}$ (+MIM)& Dynamic & 64.73±0.36 & 21.68±0.99 & 37.45±0.91 \\
 DeBERTa$_{XLARGE}^{40}$ & 7 & 63.16 & 18.57 & 34.25 \\
 DeBERTa$_{XLARGE}^{40}$ (+MIM)& 7 & 62.85±0.35 & 20.64±0.51 & 36.01±0.53 \\
 DeBERTa$_{XLARGE}^{40}$ & Dynamic & 64.24 & 17.79 & 33.8 \\
 DeBERTa$_{XLARGE}^{40}$ (+MIM)& Dynamic & 64.68±0.39 & 19.67±0.58 & 35.66±0.6 \\
 \hline
 PolyLM$_{BASE}$ \citep{ansell-etal-2021-polylm} & 8 & \textbf{64.8} & \textbf{23} & \textbf{38.3} \\
 PolyLM$_{SMALL}$ \citep{ansell-etal-2021-polylm} & 8 & 64.5 & 18.5 & 34.5 \\
 BERT+DP \citep{amrami2019better} & 7 & 64 & 21.4 & 37 \\
 LSDP \citep{amrami-goldberg-2018-word} & 7 & 57.5 & 11.3 & 25.4 \\
 AutoSense \citep{amplayo2018autosense}& Dynamic & 61.7 & 7.96 & 22.16 \\ [1ex] 
 \hline
\end{tabular}
\caption{Comparison of WSI-specific techniques on SemEval 2013 task 13}
\label{tab:semeval2013}
\end{table*}

\subsection{Experiments}
In order to prepare the training set of \textit{SemEval 2010 task 14}, we chose randomly 3500 sentences from the provided training dataset of this task for each target word. For \textit{SemEval 2013 task 13}, we extracted for each tagged target word up to 3500 random sentences from ukWac. Note that, if some of the target words in \textit{SemEval 2013 task 13} do not have 3500 sentences on ukWac, we extracted all the possible sentences. 
\begin{table}[h]
 \centering
 \small
 \begin{tabular}{||c c c||} 
 \hline
 Dataset &  Train & Test \\ [0.5ex] 
 \hline\hline
SemEval-2010 Task 14 &3.02\% & 13.5\%\\
SemEval-2013 Task 13 & 16.05\% & 9.95\% \\
\hline
\end{tabular}
\caption{The average perturbation percentage between the input text and the paraphrase. This percentage represents the proportion of changed unigrams.}
\label{tab:perturbation}
\end{table}
Following, we generate the paraphrases for both datasets by integrating the random perturbation described in section \ref{perturbation}. The average percentage of perturbation for each dataset is presented in table \ref{tab:perturbation}.
\\
The instances in \textit{SemEval-2010 task 14} and \textit{SemEval-2013 task 13} datasets contain some of the target words with morphological variability.
Hence, lemmatizing is required to identify the target lemma during the vector extraction phase. Given this word and its POS tag, we use the WordNetLemmatizer from \textit{NLTK} library to find its position inside both the sentence and its paraphrase followed by extracting the corresponding RoBERTa, BERT and DeBERTa vectors. These vectors are used to train the model as described earlier.\\
\begin{table*}[!ht]
	\centering
	 \small
	\begin{tabular}{||c||c||c||c||c||c||c||c||c||c||}
		\hline
		\multirow{2}{*}{\textbf{Model}} & \multirow{2}{*}{\textbf{\#Clusters}} & %
			\multicolumn{4}{c||}{\textbf{SemEval-2010}} &
			\multicolumn{4}{c||}{\textbf{SemEval-2013}}\\
			\cline{3-10}
			& & Layer & V-measure & F-score & AVG & Layer & F-BC & F-NMI & AVG \\ 
			\hline \hline
			RoBERTa$_{LARGE}$ & 7 & 10 & 43.6 & 68.12 & 54.5 & 9 & 63.87 & 23 & 38.32 \\
			RoBERTa$_{LARGE}$ & dynamic & 10 & 41.9 & 68.52 & 53.58 & 9 & 65.08 & 18.84 & 35.02 \\
			BERT$_{LARGE}$ & 7 & 21 & 40.8 & 66.7 & 52.17 & 20 & 63.16 & 22.07 & 37.34 \\
			BERT$_{LARGE}$ & dynamic & 21 & 41.3 & 67.65 & 52.85 & 20 & 65.54 & 21.26 & 37.32\\
			DeBERTa$_{XLARGE}$ & 7 & 32 & \textbf{49} & 69.48 & \textbf{58.35} & 33 & 64.86 & \textbf{24.14} & \textbf{39.57} \\
			DeBERTa$_{XLARGE}$ & dynamic & 32 & 46.4 & \textbf{69.49} & 56.78 & 33 & \textbf{66.62} & 21.71 & 38.03 \\
			\hline
	\end{tabular}
	\caption{The best layers of different pretrained language models on \textit{SemEval-2010 Task 14} and \textit{SemEval-2013 Task 13}}
	\label{semevalLm}
\end{table*}
\\
To infer the sense of a test instance in \textit{SemEval 2010}, we first apply the agglomerative clustering method on the extracted RoBERTa$_{LARGE}$, BERT$_{LARGE}$ and DeBERTa$_{XLARGE}$ vectors of the target word in the test instances (Section \ref{bertvectors}). The aforementioned step studies the effect of our word vectors enriching method.
Second, for the model to be tested, we forward the test word vectors to the trained model and extract the corresponding hidden state of the first layer. This state is considered as the new word representation (sense embedding) of dimension 2048.\\
Finally, we applied agglomerative clustering on the new word representations implementing our clustering solution. We assigned each instance to a single cluster.\\
The results of the evaluation on both \textit{SemEval-2010} and \textit{SemEval-2013} tasks are presented in tables \ref{tab:semeval2010} and \ref{tab:semeval2013} respectively providing the comparison with other WSI systems.\\
In the \textit{SemEval 2013} task, there is a possibility for a word to have multiple senses with a corresponding degree of applicability. Thus, once the agglomerative clustering applied, we convert the cosine similarity distances between each target word's representation and the centroids of the different clusters to a vector of probabilities using the softmax function. These probabilities are considered as the senses' degrees of applicability.
The average number of clusters for each dataset in the dynamic setting is presented in table \ref{tab:dynamic}.
    \begin{table}[h]
     \centering
     \small
     \begin{tabular}{||c c ||} 
     \hline
     Dataset &  Average \# of clusters \\ [0.5ex]
     \hline\hline
    SemEval-2010 Task 14 & 6.73\\
    SemEval-2013 Task 13 & 5.36\\
     \hline
    \end{tabular}
    \caption{The average number of clusters obtained by using the polysemy scores on SemEval 2010 and SemEval 2013 test datasets}
        \label{tab:dynamic}
    \end{table}
\begin{figure*}[h!]
    \centering
    \includegraphics[width=12.9cm]{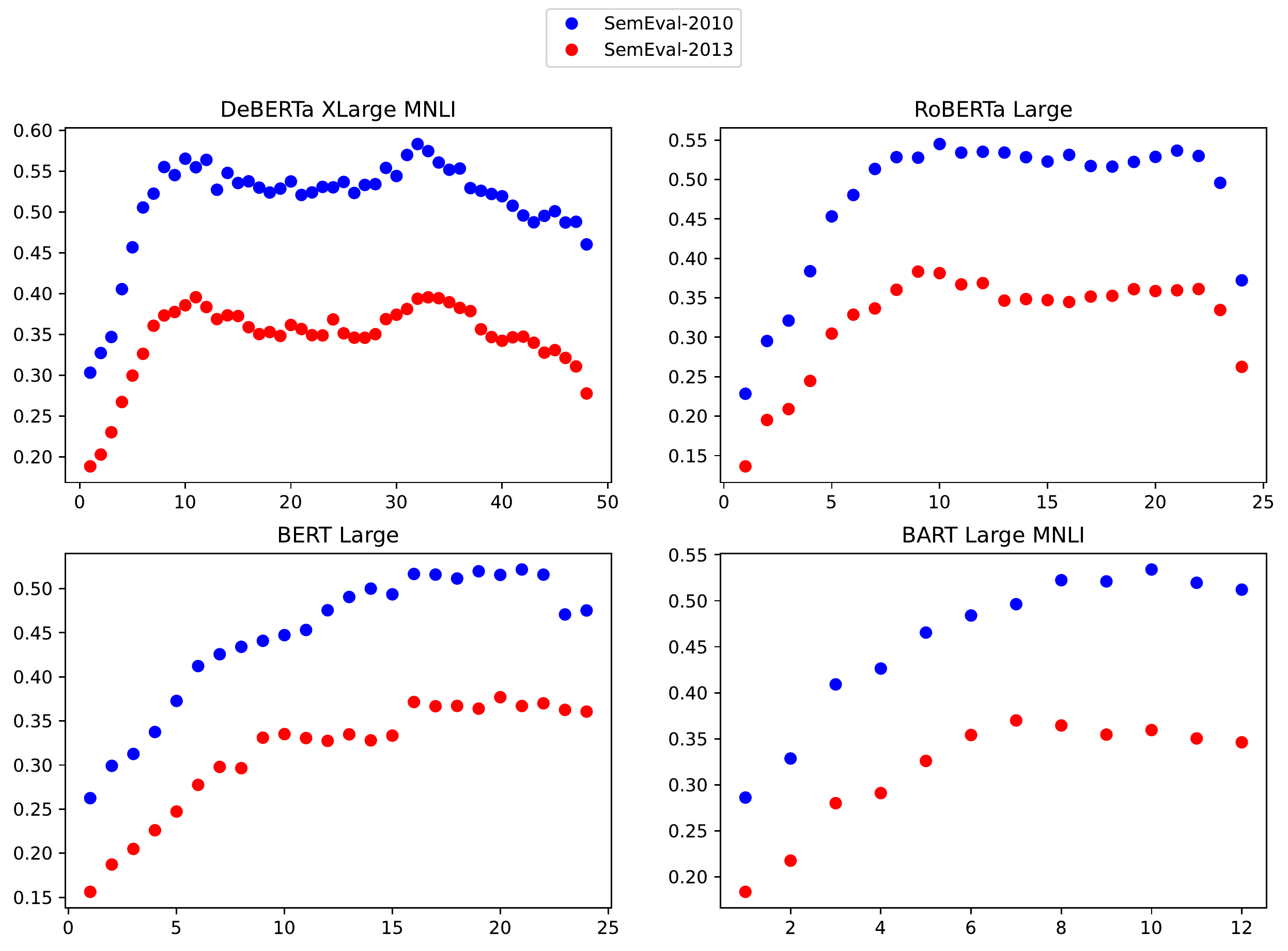}
    \caption{The AVG scores of SemEval-2010 and SemEval-2013 WSI tasks using agglomerative clustering on all the layers of different pretrained models.}
    \label{fig:layers}
\end{figure*}

\subsection{Results}
Table \ref{tab:semeval2010} shows the performance of our approach in comparison to other baselines on \textit{SemEval 2010 task 14}. The best performing system, among the baselines, is BERT+DP \cite{amrami2019better} providing the highest F-score of 71.3\%. With our method, RoBERTa$_{LARGE}$ outperforms all baselines in both settings: Fixed and dynamic number of clusters. This finding highlights the importance of our MIM approach that allows for an improvement of 1.5 absolute points over the previous state-of-the-art in terms on average score. In addition, we observe that the only model using dynamic clustering among the baselines (AutoSense) is largely outperformed by the other methods using a fixed number of clusters. However, given that WSI is an unsupervised task, the fixed number of clusters is supposed to be arbitrary and there is no guarantee that using the same number of clusters on other datasets would be optimal. Our proposed dynamic approach to choose the number of clusters did not deteriorate the performance of our method and in some cases led to a better performance (BERT$_{LARGE}^{18}$ for example). \\
\textit{SemEval 2013 task 13} performances are shown in Table \ref{tab:semeval2013}. The best performing baseline is PolyLM$_{BASE}$ providing the highest F-BC and F-NMI scores. Although our approach did not outperform this baseline, it shows to be very competitive. In fact, the results on \textit{SemEval 2013 task 13}, shows again the positive contribution of our MIM approach, as we can observe a significant improvement whenever it is applied. For example, applying MIM to RoBERTa$_{LARGE}$ with dynamic clustering led to an increase of more than 4 absolute points in terms of average score. \\
To sum things up, (1) our proposed intermediate MIM phase led on average to an improved hierarchical clustering and (2) the dynamic approach to choose the number of clusters maintained the stable and competitive performance of our different evaluated models.
\section{Best LM Layer}
During the evaluation in section \ref{sec:evaluation}, we used the list provided by BERTScore \cite{bert-score} authors regarding the best performing layer. This choice is motivated by the fact that we are dealing with an unsupervised task, thus it is not possible to tune such a hyper-parameter without access to gold annotations. However, \citet{bert-score} chose the best layer based on how good it performs in the task of machine translation evaluation. Dealing with a WSI task, there is no guarantee that the best layer is the same. Thus we carry out a study of the variation of the agglomerative clustering final score with respect to the layer used for the extraction of the vector representations. This study can help researchers in future works to choose the appropriate layer when dealing with a similar unsupervised task. \\
Figure \ref{fig:layers} shows the variation of agglomerative clustering performance in function of the depth of the chosen layer. Interestingly, we see that the variation of performance follows a similar pattern in SemEval-2010 and SemEval-2013 which can suggest a generalizable pattern over word sense induction datasets. Also, we can see that the pattern changes across different models. Despite having a similar architecture, the best layer depth in RoBERTa$_{LARGE}$ (layer 10) differs significantly with respect to that of BERT$_{LARGE}$ (layer 21). A future work should focus on this discrepancy and study the semantic information captured by each model's layers. 
Table \ref{semevalLm} presents the results regarding the best layer of each pretrained model on \textit{SemEval 2010 task 14} and \textit{SemEval 2013 task 13}. The best performing pretrained contextual embeddings for both tasks is DeBERTa$_{XLARGE}$ with a score that outperforms the state-of-the-art methods.

\section{Conclusion}
In this work, we introduced an unsupervised method for the WSI task based on the tuning of contextual word embeddings extracted from a pretrained language model. The method generates paraphrases of the input sentences. Hence, both sentences belong to the same sense cluster. Next, it uses both sentences to train a MIM neural network that maximizes the mutual information between the two sentences' outputs and minimizes the integrated match loss. The method improves on the state-of-the-art in one of the two WSI tasks.\\
We also use the polysemy score to test the dynamic number of senses setup as it claims superiority over the fixed setting in two out of six experiments. The MIM method proves, in most cases, an improvement in score while it does not deteriorate the performance in the others.\\
The extraction of representations for the target word depends on the chosen layer from the used pretrained language model. Thus, inspired by previous works, we conducts a comparison that helps the future studies in this choice. 

\section*{Limitations}
The aforementioned method presents an important improvement over some of the-state-of-art solutions for WSI tasks. However, it suffers from some limitations that are worth highlighting:\\
(1) This method is training a MIM model from scratch for each target word proving a lack of generalizability. Thus, a further study can fulfil this task by training the MIM model starting from a pretrained language model for all target words. Applying this might yield to a general model that can give the sense embedding for all possible target words before applying agglomerative clustering.\\
(2) Using the pretrained language models partially in our pipeline makes our method costly in terms of computation time when comparing with \textit{PolyLM}. As consequence, our method suffers from higher number of parameters especially with models of bigger size such as DeBERTa. Thus, a further approach is to test with smaller models (i.e DitilBERT) that could maintain the same good performance with faster training and inference time.
%

\bibliography{anthology,custom}

\begin{thebibliography}{37}
\expandafter\ifx\csname natexlab\endcsname\relax\def\natexlab#1{#1}\fi

\bibitem[{Agirre et~al.(2009)Agirre, Alfonseca, Hall, Strakova, Pasca, and
  Soroa}]{2001}
Eneko Agirre, Enrique Alfonseca, Keith Hall, Jana Strakova, Marius Pasca, and
  Aitor Soroa. 2009.
\newblock \href {https://doi.org/10.3115/1620754.1620758} {A study on
  similarity and relatedness using distributional and wordnet-based
  approaches.}
\newblock pages 19--27.

\bibitem[{Agirre and Soroa(2007)}]{agirre-soroa-2007-semeval}
Eneko Agirre and Aitor Soroa. 2007.
\newblock \href {https://www.aclweb.org/anthology/S07-1002} {{S}em{E}val-2007
  task 02: Evaluating word sense induction and discrimination systems}.
\newblock In \emph{Proceedings of the Fourth International Workshop on Semantic
  Evaluations ({S}em{E}val-2007)}, pages 7--12, Prague, Czech Republic.
  Association for Computational Linguistics.

\bibitem[{Amplayo et~al.(2018)Amplayo, won Hwang, and
  Song}]{amplayo2018autosense}
Reinald~Kim Amplayo, Seung won Hwang, and Min Song. 2018.
\newblock \href {http://arxiv.org/abs/1811.09242} {Autosense model for word
  sense induction}.

\bibitem[{Amrami and Goldberg(2018)}]{amrami-goldberg-2018-word}
Asaf Amrami and Yoav Goldberg. 2018.
\newblock \href {https://doi.org/10.18653/v1/D18-1523} {Word sense induction
  with neural bi{LM} and symmetric patterns}.
\newblock In \emph{Proceedings of the 2018 Conference on Empirical Methods in
  Natural Language Processing}, pages 4860--4867, Brussels, Belgium.
  Association for Computational Linguistics.

\bibitem[{Amrami and Goldberg(2019)}]{amrami2019better}
Asaf Amrami and Yoav Goldberg. 2019.
\newblock \href {http://arxiv.org/abs/1905.12598} {Towards better
  substitution-based word sense induction}.

\bibitem[{Ansell et~al.(2021)Ansell, Bravo-Marquez, and
  Pfahringer}]{ansell-etal-2021-polylm}
Alan Ansell, Felipe Bravo-Marquez, and Bernhard Pfahringer. 2021.
\newblock \href {https://www.aclweb.org/anthology/2021.eacl-main.45}
  {{P}oly{LM}: Learning about polysemy through language modeling}.
\newblock In \emph{Proceedings of the 16th Conference of the European Chapter
  of the Association for Computational Linguistics: Main Volume}, pages
  563--574, Online. Association for Computational Linguistics.

\bibitem[{Bagga and Baldwin(1998)}]{bagga-baldwin-1998-entity}
Amit Bagga and Breck Baldwin. 1998.
\newblock \href {https://doi.org/10.3115/980845.980859} {Entity-based
  cross-document coreferencing using the vector space model}.
\newblock In \emph{36th Annual Meeting of the Association for Computational
  Linguistics and 17th International Conference on Computational Linguistics,
  Volume 1}, pages 79--85, Montreal, Quebec, Canada. Association for
  Computational Linguistics.

\bibitem[{Baroni et~al.(2009)Baroni, Bernardini, Ferraresi, and
  Zanchetta}]{baroni2009}
Marco Baroni, Silvia Bernardini, Adriano Ferraresi, and Eros Zanchetta. 2009.
\newblock \href {https://doi.org/10.1007/s10579-009-9081-4} {The wacky wide
  web: A collection of very large linguistically processed web-crawled
  corpora}.
\newblock \emph{Language Resources and Evaluation}, 43:209--226.

\bibitem[{Chang et~al.(2014)Chang, Pei, and Chen}]{chang-etal-2014-inducing}
Baobao Chang, Wenzhe Pei, and Miaohong Chen. 2014.
\newblock \href {https://www.aclweb.org/anthology/C14-1035} {Inducing word
  sense with automatically learned hidden concepts}.
\newblock In \emph{Proceedings of {COLING} 2014, the 25th International
  Conference on Computational Linguistics: Technical Papers}, pages 355--364,
  Dublin, Ireland. Dublin City University and Association for Computational
  Linguistics.

\bibitem[{Corrêa and Amancio(2018)}]{correa2018}
Jr~Corrêa and Diego Amancio. 2018.
\newblock \href {https://doi.org/10.1016/j.physa.2019.02.032} {Word sense
  induction using word embeddings and community detection in complex networks}.
\newblock \emph{Physica A: Statistical Mechanics and its Applications}, 523.

\bibitem[{Devlin et~al.(2019)Devlin, Chang, Lee, and
  Toutanova}]{devlin2019bert}
Jacob Devlin, Ming-Wei Chang, Kenton Lee, and Kristina Toutanova. 2019.
\newblock \href {http://arxiv.org/abs/1810.04805} {Bert: Pre-training of deep
  bidirectional transformers for language understanding}.

\bibitem[{Feinerer and Hornik(2020)}]{wn3}
Ingo Feinerer and Kurt Hornik. 2020.
\newblock \href {https://CRAN.R-project.org/package=wordnet} {\emph{wordnet:
  WordNet Interface}}.
\newblock R package version 0.1-15.

\bibitem[{Fellbaum(1998)}]{wn1}
Christiane Fellbaum. 1998.
\newblock \emph{WordNet: An Electronic Lexical Database}.
\newblock Bradford Books.

\bibitem[{Goyal and Hovy(2014)}]{goyal-hovy-2014-unsupervised}
Kartik Goyal and Eduard Hovy. 2014.
\newblock \href {https://www.aclweb.org/anthology/C14-1123} {Unsupervised word
  sense induction using distributional statistics}.
\newblock In \emph{Proceedings of {COLING} 2014, the 25th International
  Conference on Computational Linguistics: Technical Papers}, pages 1302--1310,
  Dublin, Ireland. Dublin City University and Association for Computational
  Linguistics.

\bibitem[{He et~al.(2021)He, Liu, Gao, and Chen}]{he2021DeBERTa}
Pengcheng He, Xiaodong Liu, Jianfeng Gao, and Weizhu Chen. 2021.
\newblock \href {https://openreview.net/forum?id=XPZIaotutsD} {Deberta:
  Decoding-enhanced bert with disentangled attention}.
\newblock In \emph{International Conference on Learning Representations}.

\bibitem[{Hovy et~al.(2006)Hovy, Marcus, Palmer, Ramshaw, and
  Weischedel}]{ontonet}
Eduard Hovy, Mitchell Marcus, Martha Palmer, Lance Ramshaw, and Ralph
  Weischedel. 2006.
\newblock Ontonotes: The 90

\bibitem[{Ide and Suderman(2004)}]{ide-suderman-2004-american}
Nancy Ide and Keith Suderman. 2004.
\newblock \href {http://www.lrec-conf.org/proceedings/lrec2004/pdf/518.pdf}
  {The {A}merican national corpus first release}.
\newblock In \emph{Proceedings of the Fourth International Conference on
  Language Resources and Evaluation ({LREC}{'}04)}, Lisbon, Portugal. European
  Language Resources Association (ELRA).

\bibitem[{Ji et~al.(2019)Ji, Henriques, and Vedaldi}]{ji2019invariant}
Xu~Ji, João~F. Henriques, and Andrea Vedaldi. 2019.
\newblock \href {http://arxiv.org/abs/1807.06653} {Invariant information
  clustering for unsupervised image classification and segmentation}.

\bibitem[{Jurgens and Klapaftis(2013)}]{jurgens-klapaftis-2013-semeval}
David Jurgens and Ioannis Klapaftis. 2013.
\newblock \href {https://www.aclweb.org/anthology/S13-2049} {{S}em{E}val-2013
  task 13: Word sense induction for graded and non-graded senses}.
\newblock In \emph{Second Joint Conference on Lexical and Computational
  Semantics (*{SEM}), Volume 2: Proceedings of the Seventh International
  Workshop on Semantic Evaluation ({S}em{E}val 2013)}, pages 290--299, Atlanta,
  Georgia, USA. Association for Computational Linguistics.

\bibitem[{Kingma and Ba(2014)}]{https://doi.org/10.48550/arxiv.1412.6980}
Diederik~P. Kingma and Jimmy Ba. 2014.
\newblock \href {https://doi.org/10.48550/ARXIV.1412.6980} {Adam: A method for
  stochastic optimization}.

\bibitem[{Komninos and Manandhar(2016)}]{komninos2016}
Alexandros Komninos and Suresh Manandhar. 2016.
\newblock \href {https://doi.org/10.18653/v1/N16-1175} {Dependency based
  embeddings for sentence classification tasks}.
\newblock pages 1490--1500.

\bibitem[{Lau et~al.(2012)Lau, Cook, Mccarthy, Newman, and Baldwin}]{lau2013}
Jey Lau, Paul Cook, Diana Mccarthy, David Newman, and Timothy Baldwin. 2012.
\newblock Word sense induction for novel sense detection.
\newblock pages 591--601.

\bibitem[{Lewis et~al.(2020)Lewis, Liu, Goyal, Ghazvininejad, Mohamed, Levy,
  Stoyanov, and Zettlemoyer}]{lewis-etal-2020-bart}
Mike Lewis, Yinhan Liu, Naman Goyal, Marjan Ghazvininejad, Abdelrahman Mohamed,
  Omer Levy, Veselin Stoyanov, and Luke Zettlemoyer. 2020.
\newblock \href {https://doi.org/10.18653/v1/2020.acl-main.703} {{BART}:
  Denoising sequence-to-sequence pre-training for natural language generation,
  translation, and comprehension}.
\newblock In \emph{Proceedings of the 58th Annual Meeting of the Association
  for Computational Linguistics}, pages 7871--7880, Online. Association for
  Computational Linguistics.

\bibitem[{Liu et~al.(2019)Liu, Ott, Goyal, Du, Joshi, Chen, Levy, Lewis,
  Zettlemoyer, and Stoyanov}]{liu2019roberta}
Yinhan Liu, Myle Ott, Naman Goyal, Jingfei Du, Mandar Joshi, Danqi Chen, Omer
  Levy, Mike Lewis, Luke Zettlemoyer, and Veselin Stoyanov. 2019.
\newblock \href {http://arxiv.org/abs/1907.11692} {Roberta: A robustly
  optimized bert pretraining approach}.

\bibitem[{Manandhar and Klapaftis(2009)}]{manandhar-klapaftis-2009-semeval}
Suresh Manandhar and Ioannis Klapaftis. 2009.
\newblock \href {https://www.aclweb.org/anthology/W09-2419} {{S}em{E}val-2010
  task 14: Evaluation setting for word sense induction {\&} disambiguation
  systems}.
\newblock In \emph{Proceedings of the Workshop on Semantic Evaluations: Recent
  Achievements and Future Directions ({SEW}-2009)}, pages 117--122, Boulder,
  Colorado. Association for Computational Linguistics.

\bibitem[{Mikolov et~al.(2013)Mikolov, Chen, Corrado, and
  Dean}]{mikolov2013efficient}
Tomas Mikolov, Kai Chen, Greg Corrado, and Jeffrey Dean. 2013.
\newblock \href {http://arxiv.org/abs/1301.3781} {Efficient estimation of word
  representations in vector space}.

\bibitem[{Pantel and Lin(2002)}]{2002}
Patrick Pantel and Dekang Lin. 2002.
\newblock \href {https://doi.org/10.1145/775047.775138} {Discovering word
  senses from text}.
\newblock In \emph{Proceedings of the Eighth ACM SIGKDD International
  Conference on Knowledge Discovery and Data Mining}, KDD '02, page 613–619,
  New York, NY, USA. Association for Computing Machinery.

\bibitem[{Peters et~al.(2018)Peters, Neumann, Iyyer, Gardner, Clark, Lee, and
  Zettlemoyer}]{peters2018deep}
Matthew~E. Peters, Mark Neumann, Mohit Iyyer, Matt Gardner, Christopher Clark,
  Kenton Lee, and Luke Zettlemoyer. 2018.
\newblock \href {http://arxiv.org/abs/1802.05365} {Deep contextualized word
  representations}.

\bibitem[{Radford and Narasimhan(2018)}]{Radford2018ImprovingLU}
Alec Radford and Karthik Narasimhan. 2018.
\newblock Improving language understanding by generative pre-training.

\bibitem[{Rosenberg and Hirschberg(2007)}]{vmeasure}
Andrew Rosenberg and Julia Hirschberg. 2007.
\newblock V-measure: A conditional entropy-based external cluster evaluation
  measure.
\newblock pages 410--420.

\bibitem[{Song et~al.(2016)Song, Wang, Mi, and Gildea}]{song2016sense}
Linfeng Song, Zhiguo Wang, Haitao Mi, and Daniel Gildea. 2016.
\newblock \href {http://arxiv.org/abs/1606.05409} {Sense embedding learning for
  word sense induction}.

\bibitem[{Vaswani et~al.(2017)Vaswani, Shazeer, Parmar, Uszkoreit, Jones,
  Gomez, Kaiser, and Polosukhin}]{vaswani2017attention}
Ashish Vaswani, Noam Shazeer, Niki Parmar, Jakob Uszkoreit, Llion Jones,
  Aidan~N. Gomez, Lukasz Kaiser, and Illia Polosukhin. 2017.
\newblock \href {http://arxiv.org/abs/1706.03762} {Attention is all you need}.

\bibitem[{Véronis(2004)}]{article2004}
Jean Véronis. 2004.
\newblock \href {https://doi.org/10.1016/j.csl.2004.05.002} {Hyperlex: Lexical
  cartography for information retrieval}.
\newblock \emph{Computer Speech \& Language}, 18:223--252.

\bibitem[{Wallace(2007)}]{wn2}
Mike Wallace. 2007.
\newblock \href {https://sites.google.com/site/mfwallace/jawbone}
  {\emph{Jawbone Java WordNet API}}.

\bibitem[{Wang et~al.(2015)Wang, Gao, Iribarren, Lei, Xiang, Zhang, Shenghai,
  and Lu}]{wang2015}
Weicai Wang, Yang Gao, Pablo Iribarren, Yanbin Lei, Yang Xiang, Guoqing Zhang,
  Li~Shenghai, and Anxin Lu. 2015.
\newblock \emph{Wang et al. 2015}.

\bibitem[{Xypolopoulos et~al.(2020)Xypolopoulos, Tixier, and
  Vazirgiannis}]{xypolopoulos2020unsupervised}
Christos Xypolopoulos, Antoine J.~P. Tixier, and Michalis Vazirgiannis. 2020.
\newblock \href {http://arxiv.org/abs/2003.10224} {Unsupervised word polysemy
  quantification with multiresolution grids of contextual embeddings}.

\bibitem[{Zhang* et~al.(2020)Zhang*, Kishore*, Wu*, Weinberger, and
  Artzi}]{bert-score}
Tianyi Zhang*, Varsha Kishore*, Felix Wu*, Kilian~Q. Weinberger, and Yoav
  Artzi. 2020.
\newblock \href {https://openreview.net/forum?id=SkeHuCVFDr} {Bertscore:
  Evaluating text generation with bert}.
\newblock In \emph{International Conference on Learning Representations}.

\end{thebibliography}
\bibliographystyle{acl_natbib}




\end{document}